\pdfoutput=1

\documentclass[11pt]{article}

\usepackage{ACL2023}

\usepackage{times}
\usepackage{latexsym}

\usepackage[T1]{fontenc}

\usepackage[utf8]{inputenc}

\usepackage{microtype}

\usepackage{inconsolata}

\usepackage{graphicx}
\usepackage{amsmath}
\usepackage{gb4e}
\noautomath
\usepackage{booktabs}
\usepackage{lipsum}
\usepackage{color}
\usepackage{tabularx}

\usepackage[disable]{todonotes}

\makeatletter
\newcommand*\iftodonotes{\if@todonotes@disabled\expandafter\@secondoftwo\else\expandafter\@firstoftwo\fi}  
\makeatother

\newcommand{\note}[4][]{\todo[author=#2,color=#3,size=\scriptsize,fancyline,caption={},#1]{#4}} 

\newcommand{\km}[2][]{\note[#1]{km}{green!40}{#2}}

\newcommand{\mcrae}{\textsc{McRae}}
\newcommand{\buchanan}{\textsc{Buchanan}}
\newcommand{\binder}{\textsc{Binder}}

\title{A Method for Studying Semantic Construal in Grammatical Constructions with Interpretable Contextual Embedding Spaces}

  \author{Gabriella Chronis \and Kyle Mahowald \and Katrin Erk \\
    The University of Texas at Austin \\
    \texttt{\{gabriellachronis,kyle,katrin.erk\}@utexas.edu} }

\begin{document}
\maketitle
\begin{abstract}

We study semantic construal in grammatical constructions using large language models. First, we project contextual word embeddings into three interpretable semantic spaces, each defined by a different set of psycholinguistic feature norms. 
We validate these interpretable spaces and then use them to automatically derive semantic characterizations of lexical items in two grammatical constructions: nouns in subject or object position within the same sentence, and the AANN construction (e.g., `a beautiful three days'). We show that a word in subject position is interpreted as more agentive than the very same word in object position, and that the nouns in the AANN construction are interpreted as more measurement-like than when in the canonical alternation. Our method can probe the distributional meaning of syntactic constructions at a templatic level, abstracted away from specific lexemes. 

\end{abstract}

\section{Introduction}

There are now several paradigms for the linguistically oriented exploration of large neural language models. Major paradigms include treating the model as a linguistic test subject by measuring model output on test sentences \citep[e.g.,][]{linzen2016assessing,wilcox2018what,futrell-etal-2019-neural} and building (often lightweight) probing classifiers on top of embeddings, to test whether the embeddings are sensitive to  
certain properties like dependency structure \citep{tenney-etal-2019-bert,hewitt-manning-2019-structural,rogers-etal-2020-primer,belinkov-2022-probing,manning2020emergent}. \footnote{Code and data for all experiments in this paper are available at \url{https://github.com/gchronis/features_in_context}.} 

\begin{figure}[t]
    \centering
    \includegraphics[width=\linewidth]{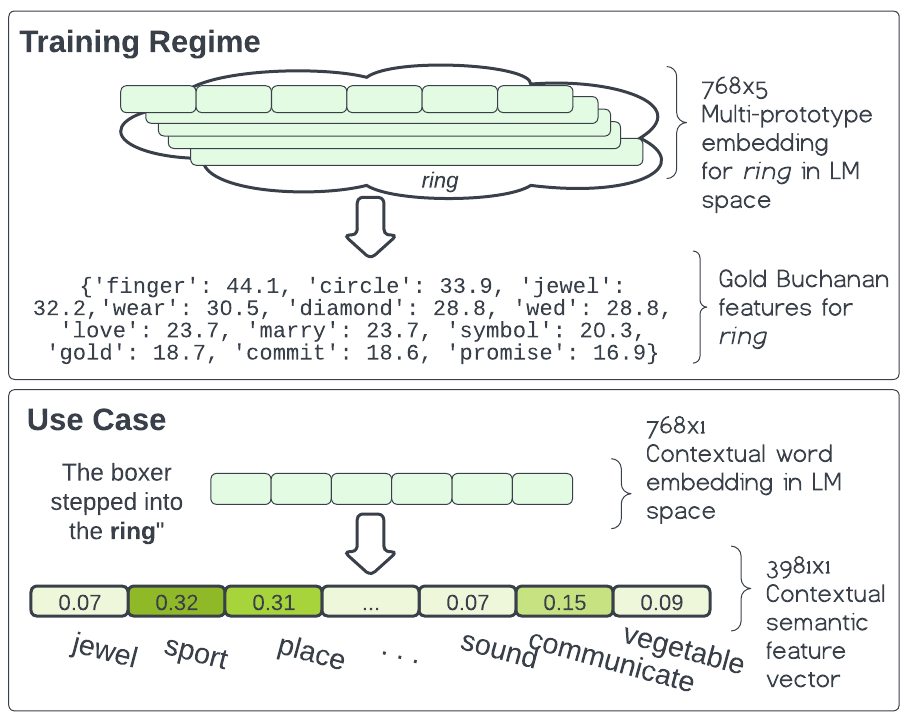}
    \caption{\textbf{(top)} Models are trained by using multi-prototype embeddings in LLM space to predict gold feature vectors derived from psycholinguistic feature norms. \textbf{(bottom)} These same models are used to project  contextual word embeddings to interpretable contextual feature space (model=\textsc{Buchanan-PLSR-MIL}).
    }
    \label{fig:features-in-context}
\end{figure}

Here, we consider another approach: projecting contextual, token-level embeddings into interpretable feature spaces defined by psycholinguistic feature norms  \cite{binderBrainbasedComponentialSemantic2016,buchananEnglishSemanticFeature2019,mcraeSemanticFeatureProduction2005}.
By learning a mapping to these spaces, as illustrated in Figure~\ref{fig:features-in-context}, we attain context-sensitive, interpretable, real-valued lexical-semantic features.

After  experimenting to determine best practices
for 
contextual-feature projection, 
we use these features to explore 
whether contextual embeddings are sensitive to subtle semantic \textit{construals} in different grammatical constructions. Specifically, we observe how even seemingly similar constructions 
can impart a different semantics on their component parts or `slot fillers' \citep{trott-etal-2020-construing,goldberg2019explain}. 
Consider the Article + Adjective + Numeral + Noun (AANN) construction: e.g., ``a beautiful three days in London,'' where the normally singular ``a'' precedes a plural noun and the adjective precedes the numeral \citep{solt_two_2007,dalrymple_amazing_2019,keenan_pleasant_2013}.
This construction often occurs with units or measure phrases (e.g., \textit{days}, \textit{feet}), but can also occur with non-measure nouns (e.g., ``a lucky three students'').

While it is tempting to think of ``a lucky three students'' as semantically equivalent to ``three lucky students,'' it has a different \textit{construal}. Specifically, the AANN construction is acceptable only when the noun behaves as a single collective unit and is, in effect, more semantically similar to a unit of measurement than it would be in the unmarked construction. Evidence for a difference in meaning between the two variants is seen in their divergent distributions.
For example, the AANN construction is unavailable in contexts like (1) and (2) \cite[\#-ed cases; adapted from][]{solt_two_2007}.

\begin{exe}
\ex \label{ex:paragraphs}{The essay consisted of (a few eloquent paragraphs / \# an eloquent few paragraphs) separated by pages of gibberish.}
\ex {He played (five boring songs / \# a boring five songs), but in between he played one really good one.}
\end{exe}

\noindent The AANN construction cannot occur in contexts where the referent of the noun is split into non-contiguous parts. This distributional pattern is taken as evidence that the AANN construction construes its argument as a single, measure-like unit. 

In this paper, we 
study distributional evidence on a larger scale, using a contextualized large language model as a `compressed corpus' that captures observed statistical regularities over utterances of many speakers. 
We analyze this compressed corpus by mapping embeddings to interpretable feature spaces based on psycholinguistic feature norms.
When we do this for the embedding of the noun \textit{days} in ``I spent a beautiful three days in London,'' we find the most salient difference with the ``I spent three beautiful \textit{days} in London'' to be 
\textbf{a higher value for features like \textit{measure} and \textit{unit} when it is in an AANN construction.}
We argue that this is because human speakers construe the AANN construction as being ``measure-ish'', and that this construal is reflected in their language use in a way that the contextual language model can pick up.

We conduct two case studies, one about AANNs and the other about grammatical subjecthood.
Specifically, \textbf{we show that a word in subject position is interpreted as more agentive than the very same word in object position} \citep[consistent with findings from psycholinguistics, e.g., ][]{kako2006thematic}, and that \textbf{a noun in the AANN construction is interpreted as more measurement-like than when in the canonical alternation.}
Our results demonstrate that construals can be inferred from statistical usage patterns. While we here use constructions with known construals, our positive results indicate that we may be able to analyze constructions where the construal is less clear in the theoretical literature.

While feature norms have been used to \textit{interpret} distributional semantic models \cite{typeDM, herbelotBuildingSharedWorld2015, fagarasanDistributionalSemanticsFeature2015, rosenfeldAnalysisPropertyInference2020}, we emphasize the \textit{linguistic} value of reliable, reusable, interpretable semantic spaces, which we use
The ability of our method to characterize 
subtle semantic differences using language models offers a point of connection between linguistically oriented deep neural network analysis \citep{baroni2021proper} and topics in formal linguistics. In particular, this work empirically demonstrates the potential alignment between LMs and feature-based theories of lexical semantics \citep[as illustrated by][]{petersen-potts-2023-lexical}.

Our main goal is to use interpretable feature spaces for understanding the semantic construal of words in context, 
specifically the AANN construction and the transitive construction. 

In Section~\ref{sec:methods}, we lay out our method for constructing interpretable feature spaces for tokens in context. Then, in Section~\ref{sec:eval}, we evaluate the success of our method on a sense differentiation task, a homonym feature prediction task,
and a qualitative analysis. The idea is that, if the method for mapping from embedding space to context-sensitive feature space is successful, we will predict unique semantic features for different senses. Having established and validated our method, we then turn to our key constructions in Section \ref{sec:constructions}.

\section{Methods}
\label{sec:methods}

The task is to learn a mapping from contextual word embedding space to an interpretable space defined by feature norms (Section \ref{sec:psych}), where every dimension corresponds to a semantic feature. We construct the training data by pairing feature norms with embeddings derived from contextual word vectors.
We train models at the type-level, e.g., to map the embedding vectors for the word  \textit{ring} to the set of feature norms for \textit{ring}, as shown in the top half of Figure~\ref{fig:features-in-context}. But ultimately, we use the model to predict semantic features for individual tokens. That is, we project the token vector of a single occurrence of the word ``ring'' into the feature space learned at the type-level, as shown in the bottom half of Figure~\ref{fig:features-in-context}. 

\subsection{Psycholinguistic feature norms}\label{sec:psych}

We construct three  semantic spaces, trained from 
three datasets of psycholinguistic feature norms. 

\textbf{The  \citet{mcraeSemanticFeatureProduction2005} feature norms} comprise 541 concrete English nouns and 2,526 features. Participants were asked to list definitional properties of cue words.
The features are full predicates; for example,  a \textit{brush} \texttt{`has\_bristles'} and is \texttt{`used\_on\_hair'}.

\textbf{ The  \citet{buchananEnglishSemanticFeature2019} feature norms} consist of over 4000 English words and 3,981 distinct features, from all open-class parts of speech, and include abstract words. The authors collect new norms and collate them with McRae norms and the \citet{vinsonSemanticFeatureProduction2008a} verb feature norms.
The features  are tokenized and lemmatized. If a participant said `found in kitchens,' this yields the features \texttt{`found'} and \texttt{`kitchen'}.

\textbf{ The  \citet{binderBrainbasedComponentialSemantic2016} data} consists of 535 English words rated for the relevance of  65 pre-defined features.
The features were chosen to correspond to known neural activation regions in the human brain, and to domains of cognition and perception; they are more coarse grained than the other norms. The word \textit{song} might have a high rating for `Audition' but a lower rating for `Vision'. 

\paragraph{Feature norms as feature spaces} Feature norms can be interpreted as vectors, with a real-valued dimension for each feature in the dataset.   
The differences between the feature norm data sets lead to differences in the  
feature inference problems.
For \mcrae\ and \buchanan, values along each feature-dimension correspond to the number of participants who named that feature---zero in the majority of cases.
These spaces are thus sparse and high-dimensional. For these two spaces, we treat the output as a ranked list of features, where the lower ranks are not relevant. 
The \binder\ space is dense and low-dimensional, and the goal is to predict the value of each feature. Here, 
a low value on a feature does not indicate lack of relevance. 

The norms differ in what they say about a word. The McRae and Buchanan norms are fine-grained, and represent salient or prototypical meanings. McRae norms are limited in their applicability because they only cover concrete nouns.
Buchanan norms have a coverage that is wider but still somewhat ad-hoc.
The Binder norms are high-level and were designed to be comprehensive.

Past and concurrent work on feature prediction has explored the utility of McRae \cite{fagarasanDistributionalSemanticsFeature2015, herbelotBuildingSharedWorld2015, rosenfeldAnalysisPropertyInference2020} and Binder \cite{utsumiExploringWhatEncoded2020, turton-etal-2021-deriving} norms for probing distributional models and language models.

\subsection{Embeddings} 
\label{sec:mpro}

The feature norms serve as our gold feature labels that we map our type-level embeddings onto. For these type-level embeddings, we use embeddings derived from BERT \cite{devlin-etal-2019-bert}, either in a \textit{vanilla} variety (one vector representation per word) or using \textit{multi-prototype embeddings}, which have multiple embedding clusters per word (roughly corresponding to distinct usages).
Specifically, we use the embeddings from \citet{chronisWhenBishopNot2020}, which are
 generated by performing K-means clustering on BERT embeddings of tokens from the British National Corpus (BNC). This procedure collects up to 200 occurrences of each cue word in the British National Corpus, and  generates token vectors for each occurrence with the HuggingFace \texttt{bert-base-uncased} model.
 For multi-prototype embeddings, these representations are clustered using K-means, using their best-performing setting of K=5 clusters per word at Layer 8. 
 For vanilla embeddings, we generate BERT vectors through the same procedure, but simply average the token vectors together (K=1) to get one vector per word.
 See Appendix~\ref{appendix:mutlipro} for more detail on the multi-prototype vectors. 

Though the mapping is \textit{trained} from type-level (or sense-level) embeddings, contextual word vectors at the token level can be \textit{projected} into the interpretable space using the resulting model.

\subsection{Mapping from embeddings to feature norms}

Though feature prediction is well explored for static embeddings \cite{typeDM, herbelotBuildingSharedWorld2015, fagarasanDistributionalSemanticsFeature2015, rosenfeldAnalysisPropertyInference2020, utsumiExploringWhatEncoded2020} and gaining popularity as a method to probe contextual embeddings \cite{chersoni-etal-2021-decoding, turton-etal-2021-deriving, apidianaki-gari-soler-2021-dolphins, proietti-etal-2022-bert}, there is no consensus as to which models work best for which datasets. We experiment with several mapping methods used previously for feature prediction. 
 The first is a feed forward neural network \citep[FFNN, with a single hidden layer, tanh activation, and dropout applied after the final output layer;][]{turtonExtrapolatingBinderStyle2020}. The dropout parameter, hidden layer size, learning rate, and number of epochs were grid-searched, as described in Appendix \ref{appendix:implementation-details} (which also includes implementation details for the other models described). The second is partial least squares regression \citep[PLSR, using the scikit-learn implementation;][]{herbelotBuildingSharedWorld2015, fagarasanDistributionalSemanticsFeature2015, utsumiExploringWhatEncoded2020}, whereby we run a partial least squares regression that predicts the feature space from the (potentially multi-prototype) embeddings. The third is label propagation \citep[PROP;][]{rosenfeldAnalysisPropertyInference2020}, which percolates labels through a graph from labels to unlabeled nodes. 
 
In all cases, the goal is to predict a real-valued semantic feature vector. Thus, the task is formulated as a multi-output regression problem.
In the vanilla setting, the above methods can straightforwardly map from a particular word embedding into feature space.
But, in order to map from a \textit{multi-prototype} embedding into feature space, the problem is trickier---especially since the multi-prototype embeddings may capture meanings that are entirely absent in interpretable feature space. 

Therefore, we test versions of each model using techniques inspired by multi-instance learning \citep[MIL;][]{dietterichSolvingMultipleInstance1997}.
The implementation of these MIL-inspired models is different for each of the three methods.
For the FFNN, we use an attention mechanism that allows the model to learn a weighted average over instances, as in \citet{ilseAttentionbasedDeepMultiple2018}.
For PLSR and Label Propagation, we simply construct a separate training example for each prototype drawn from the multi-prototype embedding
That is, for a 5-prototype vector, we construct 5 training examples, where each of the 5 examples consists of a (unique) single prototype vector paired with the same type-level feature vector.
See Appendix~\ref{appendix:mil} for more detail on adaptations for the multi-prototype setting.

\section{Evaluating Contextual Feature Norms for Interpreting Semantic Space}\label{sec:eval}

We first evaluated the models on their ability to fit the \textit{type-level} feature norms they are trained on. We do not go into detail here, as it is context-dependent meanings we are most interested in.
See Appendix~\ref{appendix:type-level-results} for full results. Overall, BERT-derived models were comparable to those we trained with static GloVe \cite{pennington2014glove} embeddings, and to the best static models in the literature.
This initial evaluation established that models using BERT-derived embeddings are just as good as static embeddings for predicting semantic features. 

To evaluate our models on \textit{in-context} feature prediction,
 we conduct two quantitative experiments: one on a sense differentiation task, one on a homonym disambiguation task, as well as a qualitative analysis for a representative word (\textit{fire}). 
The goal of this section is to explore whether the contextual feature norm method successfully captures contextual modulation of word meaning. 
For these experiments, we select the hyperparameters for each model that performed the best at type-level feature prediction under 10-fold cross-validation (Appendix~\ref{appendix:type-level-results}).

\subsection{Exp. 1: Sense Differentiation}

\begin{table}[t]
    \small
    \centering
\begin{tabular}{@{}l*{6}{>{\centering\arraybackslash}p{\widthof{MIL VANILLA}/2 - 2\tabcolsep}}@{}}
    \toprule
        & \multicolumn{2}{@{}c}{\textbf{McRae}} 
        & \multicolumn{2}{c@{}}{\textbf{Buchanan}} &
       \multicolumn{2}{c@{}}{\textbf{Binder}} \\
    \cmidrule(r){2-3} 
    \cmidrule(r){4-5}
        \cmidrule(r){6-7}

        & MIL  & Vanilla    & MIL  & Vanilla & MIL & Vanilla \\
    \cmidrule(r){2-3} 
    \cmidrule(r){4-5}
            \cmidrule(r){6-7}

    PLSR           & .41     & .39         &  .41     & .42  & .28 & .26   \\
    FFNN            & .36   & .36       &  .42   & .40 & .30 & .30   \\
    PROP             & -.03   & -.03       &  .10   & .10 & -.03 & -.03   \\ \bottomrule
\end{tabular}
\caption{Results of Sense Differentiation experiment. Pearson correlation of cosine similarities of predicted features vectors with Wu-Palmer similarity between senses. Data: pairs of  tokens of the same noun lemma in SemCor. 
\# Lemmas = 8021, \# Token-pairs = 1,045,966, p $<$ 0.0001 in all cases.
}
\label{tab:semcor-correlation-results}
\end{table}

Token-level  evaluation is tricky because there are no existing datasets for in-context feature norms.
Noting this obstacle, others utilize indirect methods like word-sense disambiguation and qualitative analysis, \cite{turtonExtrapolatingBinderStyle2020}, or forego in-context evaluation \cite{chersoni-etal-2021-decoding}.

\citet{turtonExtrapolatingBinderStyle2020} evaluate the Binder feature prediction model using the Words in Context Dataset \cite{pilehvarWiCWordinContextDataset2019}, which only labels token pairs as `same meaning' or `different meaning'.
We devise a sense differentiation experiment using the SemCor corpus, \cite{millerUsingSemanticConcordance1994}, which lets us do a more fine-grained analysis in terms of close and distant polysemy.

The logic of this experiment is that, if two senses of a word are semantically \textit{distant}, we expect the feature vectors in projected space to also be distant.
We test the quality of our predicted feature vectors by testing how well the cosine distance between vectors for polysemous words corresponds to the distance between their senses in WordNet  \cite{fellbaum_wordnet_2010}.

To build this dataset, we collect examples of noun lemmas in the SemCor corpus, 
which is annotated with WordNet senses for words in context. In SemCor, ``Water is a human right,'' is labeled \texttt{right.n.02}, \textit{an abstract idea due to a person}, while ``He walked with a heavy list to the right,'' is labeled \texttt{right.n.01}, \textit{the side to the south when facing east}.  
To counteract data imbalance, we collect only up to 30 instances of a particular word from any one WordNet sense. 
We  determine degrees of similarity between WordNet senses using Wu-Palmer similarity \cite{wuVerbSemanticsLexical1994}, which measures the degrees of separation between them.
Then, each token in the dataset is projected into interpretable semantic space.
We compute the cosine similarity between pairs of tokens and compare them to the Wu-Palmer similarity of their word senses. The key hypothesis is that we should see highly similar predicted features for tokens of the same sense, somewhat divergent features when the senses are different but related, and very different features for distant senses.

Table \ref{tab:semcor-correlation-results} shows the results. Regardless of whether we use Multi-Instance Learning, both PLSR and FFNN models show a significant correlation between the sense similarity and similarity of predicted features. We interpret this to mean that PLSR and FFNN 
reflect \textit{degree} differences of similarity between word senses.

\paragraph{Comparison to frozen BERT embeddings}
The results in Table \ref{tab:semcor-correlation-results} suggest that, at least to some extent, the projected semantic features capture information about different word senses. But to what extent?
We take it as a given that the hidden layer embeddings of \texttt{bert-base}, because they are sensitive to context, reflect differences in word senses.
Therefore, we run an additional baseline where we run the same correlational analysis using the frozen weights of \texttt{bert-base}, instead of the projected semantic feature.
That is, we compute a correlation between the cosine distance between \texttt{bert-base} vectors from Layer 8 and the WordNet-derived Wu-Palmer similarity metric.  
The correlation between cosine distance and WordNet distance for plain BERT vectors is as high as our best models (Pearson's $r = 0.41$, $p<.0001$), which suggests that, even though the feature projection method is trained on word types, our training procedure does not lead to catastrophic information loss about word \textit{tokens}. More precisely, for McRae and Buchanan datasets, PLSR learns a projection that is \textit{as contextual} as the original BERT space. Our best Binder space (FFNN) is less contextual than the original BERT space, though it still differentiates senses. This evaluation also demonstrates that Label Propagation, which is good at fitting norms at the type level \citep[as shown in Appendix~\ref{appendix:type-level-results} and][]{rosenfeldAnalysisPropertyInference2020} is not an effective method for generating contextual features.

\begin{table}[t]
    \small
    \centering
\begin{tabular}{@{}l*{4}{>{\centering\arraybackslash}p{\widthof{MIL and Vanilla and More}/2 - 2\tabcolsep}}@{}}
    \toprule
        & \multicolumn{2}{@{}c}{\textbf{McRae}} 
        & \multicolumn{2}{c@{}}{\textbf{Buchanan}}\\
    \cmidrule(r){2-3} 
    \cmidrule(r){4-5}
        & MIL  & Vanilla    & MIL  & Vanilla  \\
    \cmidrule(r){2-3} 
    \cmidrule(r){4-5}
    PLSR           & .50     & .50         &  .42     & .42     \\
    FFNN            & .50   & .50       &  .33   & .25   \\
    PROP             & .30   & .30       &  .58   & .25   \\\hline
\end{tabular}
    \caption{Results of Homonym Disambiguation Experiment. Performance on gold contextual feature prediction for homonyms (McRae and Buchanan only). Results reported are MAP@k. (n = 1093)
    }
    \label{tab:homonymous_mcrae}
\end{table}

\paragraph{Performance varies across words}

Performance on this task is not necessarily uniform across all words.
For instance, as discussed in Appendix \ref{appendix:concreteness-analysis}, performance on the sense differentiation task (using our interpretable feature projections \textit{or} the original BERT embeddings) is better for concrete words, relative to abstract words. 
We leave it to future work to further explore this, as well as other sources of heterogeneity in performance.

\subsection{Exp. 2: Homonym Disambiguation}
\label{sec:homonym-disambiguation}

The previous experiment considered many lemmas, with widely distinct as well as closely related senses. However, it is an indirect evaluation: it does not let us directly compare our projected context-dependent features to \textit{known} context-dependent feature norms.
But the \mcrae\ dataset offers a natural experiment, since it contains 20 homonymous words in disambiguated format. 
That is, separate norms exist in the \mcrae\ dataset (and per force the \buchanan\ dataset, which is a superset) for `hose (water)' and `hose (leggings)'. 
We treat these disambiguated norms as gold contextual features for tokens of these senses.
That is, we treat the \mcrae\ features for `hose (water)' as a gold label for the token ``hose'' in a sentence like ``I watered my flowers with the hose.''
As SemCor only contains a few sense-annotated tokens for each of the relevant homonyms, we use CoCA \cite{davies14BillionWord2018}, a large corpus that of largely American English news text, to collect a dataset of tokens for each homonym. See Appendix \ref{appendix:mcrae-homonyms} for details. Models were re-trained on all words in the feature norm dataset \textit{except} the held-out homonyms.\footnote{Because Binder norms do not contain any homonymous pairs, this evaluation is unavailable for \binder\ space.}

On this task, 
performance is measured as mean average precision (MAP@k) over the gold homonym features from McRae and Buchanan, where k is the number of gold features specific to each concept \cite{derby-etal-2019-feature2vec, rosenfeldAnalysisPropertyInference2020}. 
Table~\ref{tab:homonymous_mcrae} shows results. For both sets of norms, we see strong performance.
The best-performing models achieve a precision of 0.50 (on McRae) and 0.42 (on Buchanan). 
Though we cannot directly compare performance, feature prediction is generally understood to be a very hard task, with SOTA performance for static McRae feature prediction at  0.36 \cite{rosenfeldAnalysisPropertyInference2020}. This is because models will often predict plausible features that aren't in the gold feature set, like \texttt{has\_teeth} for \textit{cat} \cite{fagarasanDistributionalSemanticsFeature2015}.

\subsection{Qualitative Analysis}

\begin{table}[t]
    \footnotesize
    \centering
    \begin{tabularx}{\linewidth}{lX}
    \toprule
    Buchanan \\
    \midrule
    \textbf{1. figurative} & animal, color, light, fire, burn    \\
    \textbf{2. destructive}& destroy, build, cause, break, person        \\
    \textbf{3. artillery} & act, weapon, kill, loud, human    \\
    \textbf{4. cooking} & hot, food, wood, burn, heat   \\
    \textbf{5. N-N~compounds} & person, place, work, office, law  \\
    \midrule
    McRae \\
    \midrule
    \textbf{1. figurative} & has\_legs, is\_hard, different\_sizes, has\_4\_legs, is\_large    \\
    \textbf{2. destructive} & different\_colors, a\_mammal, made\_of\_paper, made\_of\_cement, inbeh\_-\_explodes       \\
    \textbf{3. artillery} & a\_weapon, used\_for\_killing, made\_of\_metal, is\_loud, used\_for\_war  \\
    \textbf{4. cooking} & found\_in\_kitchens, used\_for\_cooking, requires\_gas, an\_appliance, is\_hot   \\
    \textbf{5. N-N compounds} & has\_doors, used\_for\_transportation, a\_bird, has\_feathers, beh\_-\_eats \\
    \midrule
    Binder \\
    \midrule
    \textbf{1. figurative} &  Color, Needs, Harm, Cognition, Temperature   \\
    \textbf{2. destructive} &  Unpleasant, Fearful, Sad, Consequential, Harm      \\
    \textbf{3. artillery} &  UpperLimb, Communication, Social, Audition, Head  \\
    \textbf{4. cooking} &  Pleasant, Needs, Happy, Near, Temperature  \\
    \textbf{5. N-N compounds} & Biomotion, Face, Speech, Body, Unpleasant \\
    \bottomrule
    \end{tabularx}
    \caption{The most distinctive features for each prototype of \textit{fire} multi-prototype embeddings, in each of the three interpretable semantic spaces.}
        \label{tab:fire-distinctive-features}
\end{table}

In order to get a better sense of our in-context predictions, we now explore predicted features for clusters of token embeddings, extracted using the clustering procedure described in \citet{erkKatrinErkGabriella2023} (which use the same kind of multi-prototype embeddings as described in Section~\ref{sec:mpro}), for the representative word \textit{fire}. 
Focusing on a single, highly polysemous word allows us to build fine-grained intuition as to the kind of information each of our feature norms can offer.
In addition, characterizing token embedding clusters may be useful in itself: \citet{giulianelliAnalysingLexicalSemantic2020} use the term \emph{usage types} (UTs) for clusters of token embeddings, and note that they reflect word senses and other regularities such as grammatical constructions. 
UTs have proven useful for the study of semantic change. 
 However, while UTs are created automatically by clustering, researchers usually manually design labels for UTs to make their interpretation clear. An automatic labeling of token clusters with projected semantic features, as we demonstrate here, could hence be useful for studying UTs.

Our goal in this section is to take 5 UTs for the word \textit{fire} from \citet{erkKatrinErkGabriella2023} and project them into our interpretable semantic spaces (\binder, \mcrae, and \buchanan).  These UTs are: \textit{destructive} fire (e.g., ``There was a fire at Mr’s store and they called it arson.''), \textit{cooking/cozy} fire (e.g., ``They all went over to the fire for plates of meat and bread.''), \textit{artillery} fire (e.g., ``a brief burst of machine-gun fire''), and \textit{noun compounds} (e.g., ``fire brigade,'' ``fire hydrant'').
These UTs are represented as the centroids of K-means clusters of token vectors for the word \textit{fire}.

Then, we project these usage type vectors into interpretable semantic spaces, using PLSR+MIL for McRae and Buchanan, and FFNN+MIL for Binder. Predictably, the models predict similar features values in many cases, as the senses of \textit{fire} have a lot in common. For example, in \buchanan\ space, all UTs except \textit{artillery} have a high rating for `hot' (Appendix~\ref{appendix:fire-top-features}). 
To avoid this issue and get at how the usage types \textit{differ}, for each UT we average over the features predicted for the other four embedding centroids and select the features with the greatest positive difference to the target UT.
Table~\ref{tab:fire-distinctive-features} shows the features that most distinguish each UT.

The most distinctive features in Binder space are reasonable---destructive fire is indeed unpleasant, fearful, full of consequences, sad, and capable of causing harm. The \mcrae\ features are reasonable for the more concrete senses, which have synonyms that appear in the dataset (like `gun' for 3 and `oven' for 4). However, in contrast to \binder\ and \buchanan, the distinctive \mcrae\ features predicted for the more abstract UTs (1, 2, and 5) have no ready interpretation. 

\subsection{Discussion} 

\paragraph{Mapping method}  Looking at both experiments,
PLSR obtained the overall best results for predicting both Buchanan and McRae features. 
For Binder features, where the model must predict the best fit along \textit{every} dimension, FFNN does better.
Based on these experiments, we recommend using PLSR to predict definitional features like McRae and Buchanan, and FFNN to predict comprehensive features like Binder.

\paragraph{MIL} Aside from a few instances, the multi-instance framework does not drastically improve model performance. Though the positive effect is marginal, we use MIL in the case studies below.

  \paragraph{Choice of feature norms} 
The experiments above also give us insight into 
which feature space to use when. Experiment 1 shows that different senses are very distinct in McRae ($r=0.41$) and Buchanan ($r=0.41$) space, but not as distinct in Binder space  ($r=0.28$).

The qualitative look at feature predictions indicates that Buchanan and Binder models produce reasonable features for the word \textit{fire} in different contexts, including when used in a more abstract sense.  Though the best McRae model scores well overall on quantitative tasks, the qualitative analysis suggests that it does not extend well to abstract senses. This conclusion aligns with expectations, given that Buchanan and Binder norms contain features for verbs and abstract nouns, whereas the McRae norms only contains concrete nouns.

Binder feature vectors are comprehensive and good for examining abstract meanings, but Buchanan feature vectors can pinpoint more precise meanings. The case studies that follow use these feature spaces according to their strengths. To get an idea of the overarching differences between two constructions, we use \binder\ (\ref{sec:subjecthood}). To generate specific descriptions of lexical meaning in context, we use \buchanan\ (\ref{sec:aann}).

\section{Evaluating Constructions in Context}\label{sec:constructions}

Having validated that our method works for extracting meaningful, context-dependent semantic information from large language models, we turn to two target constructions: the AANN construction (described in the Introduction) and the basic transitive construction.
Crucially, in both studies, the word types are largely controlled between conditions (e.g., comparing ``The family spent a beautiful three days in London." vs. "The family spent three beautiful days in London.''), and so we compare context-dependent features derived from minimally different sentences.
This design lets us study the effect of context in a highly controlled way, without being influenced just by the identity of the words in the sentences. 

\subsection{Construction 1: `A Beautiful Three Days'}
\label{sec:aann}

\paragraph{Method}
Using a 1,000 sentence sample from \citet{mahowald-2023-discerning}'s dataset of sentences templatically constructed with varying nouns, adjectives, numerals, and templates from a variety of subtypes, we compared AANN head nouns to their equivalent ``default'' forms (e.g., ``The family spent a lovely three \textit{days} in London.'' vs. ``The family spent three lovely \textit{days} in London'').
Crucially, these form a near minimal pair.

We extracted the embeddings for the head noun token in each sentence. 
We projected the token embeddings into \buchanan\ space (using PLSR – MIL) and examined the delta between each feature, for each token, in the AANN construction vs. in the default construction.

\paragraph{Results} The top 5 features associated with the AANN construction (relative to default) were: \textbf{measure}, \textbf{one}, green, \textbf{unit}, grow. The features most associated with default (relative to AANN) were: animal, leg, child, human, please. The bolded AANN features suggest that nouns in the AANN alternation are more measure-like, and treated as more singular.
These are consistent with observations in the literature. 
Animacy-oriented words (e.g., animal, child, human) seem to be more associated with the default construction. Though this is not proposed outright in the literature, it's been observed that AANN's are more likely to be ungrammatical when the head noun is agentive \cite{solt_two_2007}.

Focusing in on a representative sentence pair that shows a particularly sharp difference, the word \textit{meals}  in ``They consumed an ugly five meals.'' is rated much higher on the \textsc{measure} (.18) and \textsc{unit} (.13) feature than the word \textit{meals} in ``They consumed five ugly meals.'' (.05 and .04, respectively).
We interpret these results as evidence that projection into the Buchanan space detects a meaningful and attested semantic difference between the AANN construction and the default construction.
Specifically, we can meaningfully detect that the construal associated with the AANN construction is more associated with measurement/units, compared to a non-AANN sentence matched on lexical content, even when the noun is not itself inherently a unit or measurement noun.

\subsection{Construction 2: Grammatical Roles}
\label{sec:subjecthood}

\begin{figure}
    \centering
    \includegraphics[width=1\linewidth]{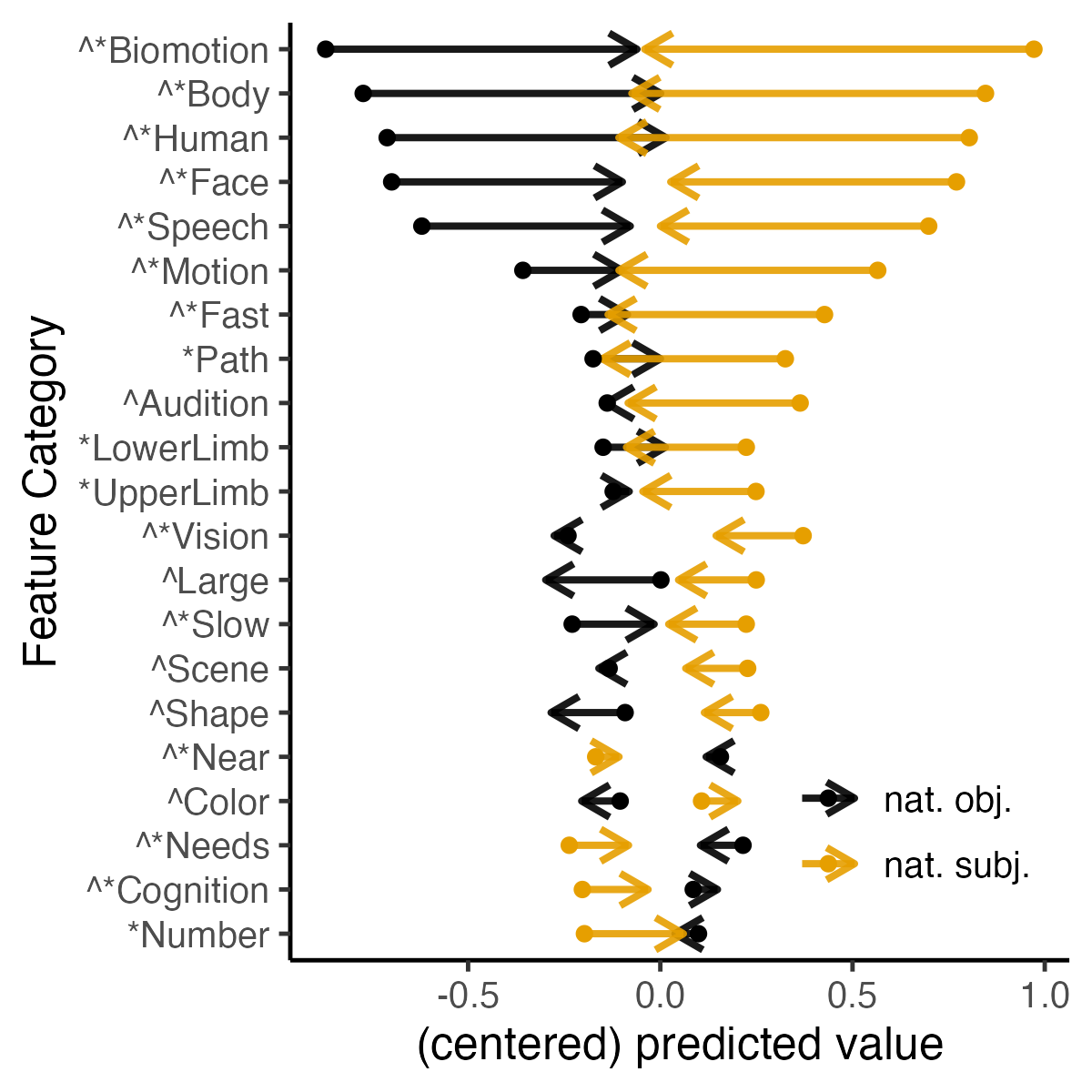}
    \caption{
    We plot the average predicted value of each feature for naturally occurring subjects and objects (points), and show how that probability shifts when we instead use  swapped sentences (arrows). We show only those features which differ significantly for either overall subjectness vs. objectness (marked with a *), or for contextual swapping (caret).
    For example, Natural Objects have low values for the Biomotion feature; when swapped to subject position, their Biomotion value increases. Norms are centered but not normalized. }
    \label{fig:subjbinder}
\end{figure}

Understanding grammatical roles like subject and object is crucial for natural language understanding. ``The dog chased the cat.'' means something different from ``The cat chased the dog.''
English relies largely on SVO word order for discriminating subjects vs. objects.
Arguments that are animate, sentient, cause an event or a change of state in another participant, or move relative to another participant tend to be realized as subjects. 
Arguments that undergo a change of state, or are affected by another participant, tend to be realized as objects \citep{levin2005argument,dowty1991thematic}.
Most of the time, just knowing the two nouns in a transitive sentence is enough to know which is the subject and which is the object: If the nouns are ``dog'' and ``bone'', you can guess that ``dog'' is the subject and ``bone'' the object~\citep{mahowald2022grammatical}.

There is evidence that contextual language models like BERT represent subjecthood \citep{linzen2016assessing,papadimitriou-etal-2021-deep,hewitt-manning-2019-structural}.
But do these models actually represent abstract grammatical subject, or do they rely on lexical information?
One way to tease this apart is to study sentences where grammatical context and lexical heuristics come apart. 
\citet{papadimitriou-etal-2022-classifying-grammatical} showed that BERT can reliably distinguish between grammatical subject and object, even for sentences with non-prototypical arguments like, ``The onion chopped the chef'', but only in the higher levels of the model after more information has been shared.
At lower layers, the model seems to rely on lexical information (e.g., would classify ``chef'' as the subject and ``onion'' as the object).

While prior work has explored the subject/object classification question by training bespoke probes, here we use projections into \textsc{Binder} space.
We focus on the set of English sentences studied in \citet{papadimitriou-etal-2022-classifying-grammatical}, which are extracted from the Universal Dependencies Treebank \citep{nivre2016universal} and appear in two forms: the original form and a form in which the subject and object are swapped. 
For instance: compare the \textsc{Natural}, ``Finally a chambermaid stuck her head around the corner'' vs. the \textsc{Swapped}, ``Finally a head stuck her chambermaid around the corner.'' The Treebank from which the sentences are sampled contains data from a number of different English corpora.

We project the subject and object in each of the 486 \textsc{Natural} sentences into \binder\ space, using the FFNN-MIL method (which is best for token-level \binder\ prediction), and then do the same for each of their \textsc{Swapped} counterparts.
We first ask whether naturally occurring subjects tend to be more animate than objects. But we then ask whether, merely by virtue of being a subject, the lexical item takes on a more animate construal.
Such a result would be consistent with psycholinguistic findings in humans: \citet{kako2006thematic} shows that, even with nonce sentences like ``The rom mecked the zarg,'' the subject word ``rom'' is rated as more animate.

\paragraph{Words that tend to appear in subject position are associated with higher animacy ratings.} Given that there are known to be systematic differences between subjects and objects, will the Binder features for subjects and objects systematically differ 
in the \textsc{Natural} sentences?
As can be seen in Figure~\ref{fig:subjbinder}, the answer is clearly yes. Animacy-associated features like Biomotion, Body, and Human are higher for naturally occurring subjects than for objects.
We ran a linear regression predicting the Binder value from the subject/object status of the word, the Binder feature, and their interaction.
The interaction term is the one we care about: how does the predicted value for that feature change when we are dealing with a subject or object? 
After Bonferroni correction for multiple comparisons, we find several features significantly correlated with subjecthood and a few with objecthood, starred in Figure~\ref{fig:subjbinder}.

\paragraph{The \textit{same token} is construed as more animate when it appears in subject position.} The preceding analysis could have been done using type-level Binder features: the upshot is that word \textit{types} that appear in subject position get animacy-associated features.
The highest rated words in this data set, for the Biomotion category, are: \textit{animals}, \textit{reptiles}, \textit{cat}, \textit{dog}, and they all occur as subjects in the corpus.
But merely knowing that naturally occurring subjects and objects differ in Binder features does not tell us the whole story. 
Using the contextual feature projections, we can explore whether two tokens of the same type are construed as differing in animacy, based on whether they appear as a subject.
We can do this in a controlled way by comparing the same word in the natural sentences and the swapped ones. For instance, in the sentence above, “chambermaid'' appears as a subject but is an object in the swapped version. 
How does its Binder rating change? 
To assess that, we compare natural subjects vs. those same words moved to object position of the same verb in the same sentence. And we compare natural objects to those same words swapped to be subjects. 
Figure~\ref{fig:subjbinder} shows that subject-oriented features like Biomotion, Body, and Human lose their large values and become more neutral.
The careted features in the figure show significant effects of being swapped, after Bonferroni correction.

To assess whether our contextual feature predictions are sufficient for predicting whether a noun is a subject, no matter if natural or swapped, we run a forward-stepwise logistic regression on a portion of the data (300 sentences) to predict whether a particular token is a subject or an object based on its Binder ratings. The forward-stepwise part picks the set of Binder features that give the best prediction. 
We then test its k-fold cross-validation accuracy on the held-out test set.
For \textsc{Natural} sentences, this method achieves 80\% accuracy, compared to 73\% accuracy for \textsc{Swapped} sentences.
Thus, while natural sentences are easier, even the swapped sentences can be  categorized better than chance using the feature norms---despite the fact that the words in question naturally occurred in the opposite roles.

We then performed the same procedure, but instead predicted whether a particular token was from a \textsc{Natural} or \textsc{Swapped} sentence. 
We did this separately for subjects and objects.
Performance was above chance, at 70\% and 71\% respectively.

So a model can, with better than chance accuracy, use projected Binder features to identify which nouns are subjects in swapped sentences.
But we can also predict which nouns are from swapped sentences.
This result suggests that the predicted Binder features reflect contextual information, but also retain type-level information. 

The results of our study align  with \citet{gianlucalebaniInvestigatingDowty82172021}  who investigate semantic proto-roles using distributional models and with  \citet{proietti-etal-2022-bert}, who investigate semantic proto-roles by projecting BERT into an interpretable space (similar to our method). Both show that transitive verbs have more proto-agent properties than their intransitive counterparts. The present analysis confirms and expands on their finding that BERT captures semantic role information and that projecting into interpretable space is a fruitful way of gaining insight into grammatical and thematic roles.

\section{Conclusion}

In this paper, we honed techniques for predicting
semantic features for token embeddings. These projections are versatile. Once created, one and the same model can be used to study a wide array of phenomena. We explored their utility for studying semantic construal in syntactic constructions. 
We emphasize the potential of this method to answer linguistic questions about meaning differences in constructions that are less well-understood and well-theorized than the ones studied here.
As such, we hope it will be possible to use this method to generate linguistic insight.
\section*{Limitations}

One limitation of our study is that interpretable feature spaces are at times only semi-interpretable.
We infer from patterns of model behavior that Buchanan features such as `human’, `child’, and `animal’ can be signals for animacy more broadly construed.
The need to conjecture about what a feature means  points to a weakness in our approach. Some interpretation will always be necessary, and with a more heavy-handed probing method like ours, it can't be certain what effects are coming from the model and which are coming from the probe.

One way to get around this need for subjective interpretation is to train a separate classifier for animacy more broadly understood, and then use the feature prediction model to examine what features are most relevant to the classifier \cite{chersoni-etal-2021-decoding}. However, this method is not foolproof either.  The classification distinction is wholly determined by the labeled data used to train the animacy probe, and the judgments are subjective. Even for a seemingly straightforward feature, the correct label is not always clear. Is a clock that \textit{sings} the hour animate? What about a \textit{stony face}?

Subjective interpretation is an important and unavoidable component of both linguistic and neural language model analysis. The goal of data-driven research is to extend the sphere of concern beyond self-reflexive subjective judgments of the researcher to the shared subjectivities of a language community. Information about animacy reflected in an annotated dataset still reflects subjectivities, but shared ones. It is important to always be clear about where interpretation is happening, whose interpretations are taken into account, and how they affect what conclusions may be drawn.

On that note, there are a few places where design decisions affect our analysis of lexical variation. Linguistic data enters the modeling pipeline in at  least four places: BooksCorpus and Wikipedia data used to pre-train BERT, the BNC corpus which we use to derive multi-prototype embeddings, the feature norm datasets which tend to capture the subjectivities of American college students, and the texts we analyze in our case studies (both natural language text and constructed examples). These resources all cover English, but necessarily reflect different varieties of English, given that they were collected in different places at different times. For example, usage types in the BNC often differ from those derived from Wikipedia data. 

Not only do the corpora we use represent potentially disjoint varieties (English spoken by college students in Vermont, English in newswire and fiction genres, English in reference texts). They also all represent the semantics of the unmarked,  \textit{normative varieties} of English. Normative English  dominates all data collection contexts upon which our study rests. Consequently, to the extent that our model is a proxy for English semantic judgments, it is a proxy for dominant semantic associations among the composers of these texts and participants in the feature norm studies. 

Though it is interesting and useful to study the English language as a whole, care must be taken to ensure that the sample is representative of all speakers; and
ideally, our approach supports linguistic approaches which aim to describe and explain the semantics of smaller language communities. This would require language models trained on corpora at the level of communities of practice, as well as feature norms specific to these communities. We are hopeful that the future of statistical methods in lexical semantic analysis moves in this direction.

\section*{Ethics Statement}

Our models are developed and published in order to encourage academic research in descriptive linguistics.  In the future, we plan to use our method to study the inherent non-neutrality of language models by examining the influence of training corpus composition on the semantic representation of social meanings, as represented by cultural keywords. Because they are built on top of an unpredictable language model, the feature prediction methods, as well as the models we publish, are recommended for descriptive research only. Researchers should take into account the potential for language models, like language, to reflect of harmful ideologies such as sexism, racism, homophobia, and other forms of bigotry.

\section*{Acknowledgements}

This work was made possible through funding from an NSF GRFP Grant to GC, NSF Grant 2139005 to KM. Thank you to the UT Austin Linguistics Computational Linguistics group for helpful comments and the SynSem group for their enthusiasm in considering how language modeling might inform their questions in semantics. For helpful discussions, thanks to Adele Goldberg and the Princeton language group, Richard Futrell, and Isabel Papadimitriou.

\bibliography{anthology,QP,custom,aann}
\bibliographystyle{acl_natbib}

\appendix


\section{Embedding Details}
\label{appendix:mutlipro}

For training, we use the multi-prototype embeddings of \citet{chronisWhenBishopNot2020}. They are generated by performing k-means clustering on BERT embeddings of tokens from the British National Corpus (BNC). This procedure collects up to 200 occurrences of each cue word in the BNC and generates tokens vectors for each occurrence with the HuggingFace \texttt{bert-base-uncased} model. These representations are then clustered using K-means, using the authors' best-performing setting of K=5 clusters per word at layer 8. 
These multi-prototype vectors are unordered, `bag-of-senses' representations.

For the static embedding baseline, we use the pretrained Wikipedia 2014 + Gigaword 5  pretrained GloVe with 300 dimensions, which is trained on 6B tokens with 400K vocabulary word \cite{pennington2014glove}.

For token-level evaluations in Section~\ref{sec:eval} above, it does not make sense to compare to GloVe because GloVe embedding space is not contextual. Instead, we compare the multi-prototype, MIL models to single prototype (vanilla) versions of each model.
Embeddings for the vanilla models are generated using the same procedure described above for multi-prototype, but all tokens are averaged into a single vector representation (K=1) rather than clustering them into prototypes.

\section{Model Implementation Details}
\label{appendix:implementation-details}

For all models, we train using ten-fold cross-validation with an 80-10-10 train-dev-test split. For the MIL models, no prototypes of the same word are repeated between train and test sets. For each prediction task, we tune model hyperparameters using a sampled grid search (see uploaded code and data for details). The chosen hyperparameter settings are the ones with the best average performance on the dev set across folds. 

The FFNN model is implemented in PyTorch and trained using the Adam optimizer with stochastic gradient descent. We search over number of epochs (30, 50); dropout (0, .2, .5), learning rates (1e-5, 1e-4, 1e-3), and hidden layer size (50, 100, 300).

Partial Least Squares regression is a statistical method to find the fundamental relations between two matrices (semantic spaces). PLSR is useful in this case because it allows for correlations among the independent variables (embedding dimensions). We use the PLSR implementation from scikit-learn. We grid search over the number of PLSR dimensionality components (50, 100, 300). 

Label propagation uses code from \citet[][]{rosenfeldAnalysisPropertyInference2020}. Models were trained on a 2.3 GHz 8-Core Intel Core i9 processor with 16 GB of RAM. In label propagation, each labeled training example is embedded as a node in a graph along with unlabeled training data. Training takes place iteratively; in each iteration, labels spread through the graph. In this method, word embeddings are labeled with their corresponding features, withholding labels from the test set. Unlabeled nodes receive features of labeled nodes which are nearby in embedding space. \citet{johnsPerceptualInferenceGlobal2012} first applied this method to feature prediction from distributional models. In their model, the features of an unlabeled word are calculated as a weighted sum of the feature values that labeled words have---the weights are determined by cosine distance in distributional semantic space. \citet{rosenfeldAnalysisPropertyInference2020} evaluate more sophisticated approaches to label propagation, called modified absorption. With modified absorption, labels do not propagate under certain conditions. For instance, features won't propagate to words that are very unfamiliar, or to words which are already well-labeled with properties.

\section{Predicting with Multi-Prototype Embeddings}
\label{appendix:mil}

The classic MIL problem is a classification task. The input is an unordered bag of instances, and the output is a binary classification label. The label of the whole bag is 1 if at least one of the instances in the bag has the label 1. However, the labels of the individual instances are unknown---only the bag labels are available. 
We take this as inspiration for our scenario, where we have a multi-prototype representation, along with a feature vector that may reflect only one of the prototypes (as in the \textit{ring} example above).

To make the FFNN suitable for MIL, the FFNN is extended by an attention mechanism without ordering, as in \citet{ilseAttentionbasedDeepMultiple2018}.
This method computes a weighted average over the instances. Code for the attention module was adapted from their implementation, and can be found at \url{https://github.com/AMLab-Amsterdam/AttentionDeepMIL}. It was used in combination with the attention module defined in this blog post: \url{https://medium.com/swlh/multiple-instance-learning-c49bd21f5620}.

To adapt PLSR for MIL, we construct one training example for each prototype.  That is, for a 5-prototype vector, we construct 5 training examples, one for each vector, labeled with the type-level features. Thus, we conduct PLSR on a dataset with noisy labels. No prototypes of the same word are repeated between train and test sets.

Similar to PLSR, to adapt Label Propagation for multi-prototype embedding inputs, we represent each prototype as an independent node that maps to a type-level feature vector.

\section{Type-level Evaluation Results}
\label{appendix:type-level-results}

Results are reported on the type-level training task. These evaluations show how well the different models are able to fit the different feature norms. We find that all models are on par with the performance reported in the existing literature on inferring static semantic features \cite{fagarasanDistributionalSemanticsFeature2015, herbelotBuildingSharedWorld2015,derby-etal-2019-feature2vec}. 

Our goal is to predict semantic feature norms from words in context.  We define a mapping problem from contextual-language-model-derived embeddings to an interpretable semantic space defined by psycholinguistic feature norms.
The training data are experimentally collected semantic features for word \textit{types}. Each consists of a cue word and a feature vector. We compare MIL and vanilla versions of FFNN, PLSR, and Label Propagation models.

The literature on feature prediction uses different evaluation methods. For \mcrae\ and \buchanan\ prediction, where the goal is to produce the most important features, we report Mean Average Precision at K (MAP@K), where K is the number of gold features for a concept \cite{derby-etal-2019-feature2vec}. For Binder vectors, every feature is valued for every word, MAP@k is always equal to 1. For \binder, where the goal is to capture the relative importance of each feature, precision is not an appropriate metric.
In this case, we use mean squared error (MSE) to measure the best overall fit.

\begin{table}
\small
\centering
\begin{tabular}{llll}
\toprule
Model   &   \mcrae  &   \buchanan & \binder \\
    &   MAP@k ($\uparrow$)      &   MAP@k ($\uparrow$)    &  MSE ($\downarrow$)\\
\midrule
\multicolumn{2}{l}{PLSR} \\
\midrule
BERT MIL &   0.33       &    \textbf{0.37}   & 2.32 \\
BERT Vanilla & \textbf{0.34}  & 0.29  & 2.37                          \\
GloVe & 0.33  & 0.23 & 2.37                            \\
\midrule
\multicolumn{2}{l}{FFNN} \\
\midrule
BERT MIL &   0.32       &   0.26   & 0.82  \\BERT Vanilla & 0.32    & 0.26   & 0.88      \\       GLoVe & 0.30  & 0.26 &  1.14      \\
\midrule
\multicolumn{2}{l}{PROP} \\
\midrule
BERT MIL &    0.31      &     0.32 & 0.96   \\
BERT Vanilla & 0.32   & 0.30    & \textbf{0.10}      \\
GloVe  & 0.30      & 0.26     &  0.89\\      
\bottomrule
\end{tabular}

\caption{Type-level performance of models trained with BERT-derived and GloVe embeddings on \mcrae\,  \buchanan\, and \binder\ feature norm prediction tasks. Bolded cells indicate the highest-performing models for each feature prediction task.}
\label{tab:type-level-results}
\end{table}

Performance overall matched the best results in the literature for static feature prediction, and models that used the BERT embeddings performed as well or better compared to training on static GloVe embeddings (Table~\ref{tab:type-level-results}). 
On the \mcrae\ prediction task, PLSR and label propagation perform the best, but the scores are more or less similar across the board. The best performance was within range of the best MAP@k scores reported in the literature \cite[MAP@k = .36 on \mcrae, per][]{rosenfeldAnalysisPropertyInference2020}. BERT embeddings produce features comparable in performance to GloVe vectors. For \buchanan, BERT models do not improve over GloVe vectors.
MIL did not fare any better than single-instance learning at the type level, with the exception of PLSR for \buchanan\, which led to a large performance gain. 

These results confirm the finding of \citet{rosenfeldAnalysisPropertyInference2020} that Label Propagation with modified absorption does very well at the task of feature prediction (or property inference, as they call it).  

However, as described in the main text, our implementation of Label Propagation is not good at modeling context-sensitive lexical-semantic phenomena unless it is supplied with unlabeled nodes for different senses at training time. 
Label Propagation under the MIL condition did a particularly good job at disambiguating homonyms (Table~\ref{tab:homonymous_mcrae}), provided that the different senses were given as unlabeled nodes during training. 
However, Label Propagation does very poorly on the sense differentiation task (Table~\ref{tab:semcor-correlation-results}), showing that this model does not predict different features for different senses when it is not exposed to unlabeled nodes for these senses during training. We believe this is a consequence of the number of nodes in our graph.
At test time, PROP is limited to a fixed number of potential features---given any context vector, it retrieves the closest vector in the graph and gives those labels. Unless there are very many nodes in the graph for each word, PROP will often return the same features for different senses, because there is a high pairwise similarity in BERT space among tokens of the same type \cite{mickus-etal-2020-mean}.

Performance for Label Propagation should improve with the number of unlabeled nodes included during training, but this increases runtime and is not feasible for large datasets or convenient for \textit{ad hoc} linguistic analyses like those we wish to apply the feature prediction model to.

\section{Concreteness Analysis}
\label{appendix:concreteness-analysis}

\begin{figure}
    \centering
    \includegraphics[width=\linewidth]{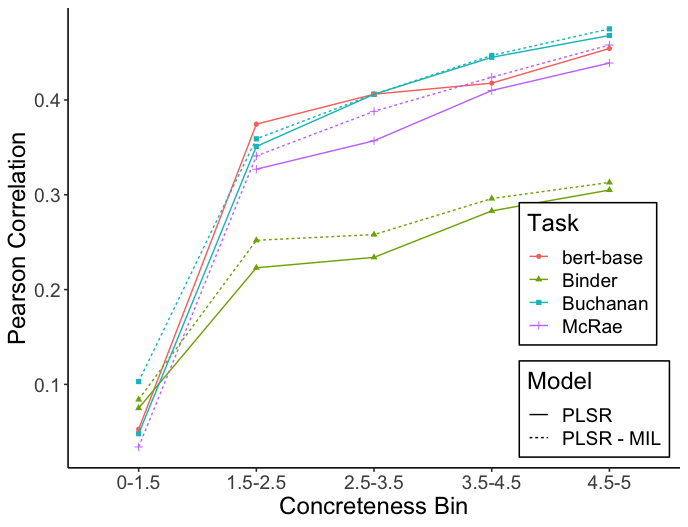}
    \caption{Pearson correlation of cosine similarities between predicted feature vectors with Wu-Palmer similarity between senses. Data: pairs of tokens of the same lemma in SemCor, broken down by lemma concreteness according to Brysbaert.  
    }
    \label{fig:semcor-correlation-by-concreteness}
\end{figure}

\begin{table*}
    \small
    \centering
    \begin{tabular}{lllll}
    \toprule
    Usage Type 1        & Usage Type 2     & Usage Type 3   & Usage Type 4 & Usage Type 5       \\
    (transformative) & (destructive) & (artillery) & (cooking) & (N-N compounds) \\
\midrule
    fire             & fire          & act         & fire      & fire            \\
    hot              & hot           & fire        & hot       & person          \\
    burn             & burn          & danger      & burn      & hot             \\
    light            & light         & kill        & light     & burn            \\
    danger           & destroy       & weapon      & heat      & light           \\
    heat             & danger        & human       & wood      & red             \\
    cook             & heat          & metal       & cook      & danger          \\
    red              & cook          & loud        & danger    & heat            \\
    wood             & hurt          & light       & warm      & cook            \\
    act              & red           & hurt        & food      & destroy        \\
    \bottomrule
    \end{tabular}
    \caption{Top 10 predicted Buchanan features for the centroids of 5-k-means clustering on BERT token embeddings from the BNC. (model = PLSR - MIL)}
    \label{tab:fire-cluster-features}
\end{table*}

\begin{table*}
\small
\centering
\begin{tabular}{lp{7.5cm}p{5.5cm}}
\toprule
Homonym & Sentence  &  Gold Feature Norms \\
\midrule
bat (animal) & I was particularly surprised to see a tame golden fruit bat, hanging upside down on a tree branch in the morning sunshine. & wing, fly, nocturnal, black, cave, fur, animal \\
\midrule
bat (baseball) & I was at the plate. He threw; I swung the bat. The ball rocketed into left field. & hit, wood, ball, metal, long, sport \\
\bottomrule
\end{tabular}
    \caption{Example data for the Buchanan homonym disambiguation task. Sentences from COCA containing homonyms are paired with a feature norm that targets the disambiguated sense.}
    \label{tab:homonym-features}
\end{table*}

In all spaces\km{if not in main text, maybe leave this out?}, concrete polysemous senses are more clearly separated than abstract senses. This is shown in Figure~\ref{fig:semcor-correlation-by-concreteness}, which breaks down sense differentiation results by their concreteness ratings according to \citet{Brysbaert}. The problem is worst for McRae, and least pronounced for Binder. This may be due to even more variation in meaning for abstract words, which tend to be highly polysemous. Indeed, the same pattern is observed in the frozen BERT space: for concrete words, cosine similarity of token vectors is not strongly correlated with WordNet distance. 

Qualitative examination of predicted features reveals that the models are not bad at abstract meanings. For example, consider the sentence ``People travel many miles to gaze upon this natural wonder, though few are willing to approach it closely, since it is reputed to be the haunt of various demons and devils.'' 
Our Buchanan model predicts plausible features for the rather abstract `haunt': `one', `face', `dead', `bad', `body', `place', `person'. 
But the McRae model, which did not see abstract words in training and whose features only cover very concrete nouns, does not produce plausible features: `is\_expensive', `is\_smelly', 'made\_of\_wood', `is\_large'. Predicted Binder features are also plausible: `Vision', `Harm', `Unpleasant', `Sad', `Consequential', `Attention', `Angry'.

This analysis does not reflect model performance on abstract words so much as it points to a potentially interesting relationship between abstract words in BERT space and in WordNet. Do contextual vectors primarily reflect different kinds of meaning for abstract words besides word sense?

\section{Top predicted features for sense clusters}
\label{appendix:fire-top-features}

Table~\ref{tab:fire-cluster-features} shows the top 10 Buchanan features for each centroid of the usage type clusters for \textit{fire} ($k$=5, tokens taken from the BNC). Many of the most salient features are the same across the different usage types. Meanings specific to each sense and usage type are more evident when one focuses on the most \textit{distinctive} features for each cluster (Table~\ref{tab:fire-distinctive-features}).

\section{McRae Homonym Dataset Collection Procedure}
\label{appendix:mcrae-homonyms}

We train our contextual model at the type level because of the present lack of in-context feature norms to use for training and evaluation.  To evaluate at the token level directly, as described in Section~\ref{sec:eval}, we use the features that \citet{mcraeSemanticFeatureProduction2005} collected for disambiguated homonyms.

For this evaluation, we construct a test set of sentences containing these homonyms, each labeled with the feature vector for that homonym. SemCor, the sense-annotated dataset used for the sense-differentiation evaluation, does not contain enough tokens of each of the homonyms. So, we turned to the Corpus of Contemporary American English \cite{davies14BillionWord2018}. The data were collected using the following procedure:

For each homonym,
(1) Search for the target word.
(2) Read through a random sample of occurrences of the word, highlighting sentences that unambiguously use the target sense.
(3) The same researcher double-checks the list to filter out accidental sense mismatches.

At least 20 tokens of each homonym were collected, stopping at 50 (with an average of 40 contexts per sense). Table~\ref{tab:homonym-features} shows two examples from the resulting dataset. The list of homonyms and the number of tokens for each one is given in Table~\ref{tab:mcrae-homonym-list}, and the full dataset is available in the supplemental data. 

\begin{table}[ht]
\footnotesize
\centering
\begin{tabular}{lll}
\toprule
Word    & Sense     & \# Tokens \\
\midrule
bat                   & animal    &     52 \\
bat                   & baseball  &     51 \\
board                 & black     &   28   \\
board                 & wood      &   56   \\
bow                   & ribbon    &     43 \\
bow                   & weapon    &     52 \\
cap                   & bottle    &     20 \\
cap                   & hat       &    207 \\
crane                 & animal    &   14   \\
crane                 & machine   &  101   \\
hose                  & tube   & 42                   \\
hose                  & leggings   &   55  \\
mink              & animal & 32                   \\
mink                  & coat    &      33  \\
mouse            & animal &  64                   \\
mouse                 & computer  &   78   \\
pipe                  & plumbing  &    27  \\
pipe                  & smoking   &    20  \\
tank                  & army      &    35  \\
tank                  & container &    83  \\
\bottomrule
\end{tabular}
\caption{List of cue words used in homonym disambiguation experiment along with the number of tokens of each homonym collected from CoCA for the dataset.}
\label{tab:mcrae-homonym-list}
\end{table}

\section{Licenses}
\label{appendix:licenses}

{\small
    \begin{tabular}{lp{3cm}}
    \toprule
        Dataset/Model & License \\
    \midrule
    McRae Feature Norms & unknown \\
       Buchanan Feature Norms  & GPL 3.0 \\
       Binder Feature Norms  & CC BY-NC-ND 4.0 \\
       Multi-Prototype Embeddings & CC BY-NC 4.0 \\
       BNC  &  \url{http://www.natcorp.ox.ac.uk/docs/licence.html} \\
       \texttt{bert-base-uncased} & Apache 2.0 \\
       SemCor & Apache 2.0 \\
       Brysbaert Concreteness Norms &  CC BY-NC-ND 3.0 \\
       AANN Sentences & CC BY-NC-ND 4.0 \\
    \bottomrule
    \end{tabular}
    }

\end{document}